\documentclass[sigconf]{aamas}

\usepackage{balance} 

\usepackage{graphicx}
\usepackage{amsmath} 
\usepackage{algorithm}  
\usepackage{algorithmicx}
\usepackage{algpseudocode} 
\usepackage{graphicx}
\usepackage{multirow}
\usepackage{subfigure}
\usepackage{comment}
\usepackage{booktabs}
\usepackage{stackengine}
\usepackage{xcolor,colortbl}
\usepackage{amsthm}     
\usepackage{caption}
\usepackage{wrapfig}
\usepackage{arydshln}
\usepackage[most]{tcolorbox}
\usepackage{lipsum}
\usepackage{color, colortbl}

\definecolor{Gray}{gray}{0.85}

\makeatletter
\newenvironment{breakablealgorithm}
  {
   \begin{center}
     \refstepcounter{algorithm}
     \hrule height.8pt depth0pt \kern2pt
     \renewcommand{\caption}[2][\relax]{
       {\raggedright\textbf{\ALG@name~\thealgorithm} ##2\par}%
       \ifx\relax##1\relax 
         \addcontentsline{loa}{algorithm}{\protect\numberline{\thealgorithm}##2}%
       \else 
         \addcontentsline{loa}{algorithm}{\protect\numberline{\thealgorithm}##1}%
       \fi
       \kern2pt\hrule\kern2pt
     }
  }{
     \kern2pt\hrule\relax
   \end{center}
  }
\makeatother

\setcopyright{ifaamas}
\acmConference[AAMAS '26]{Proc.\@ of the 25th International Conference
on Autonomous Agents and Multiagent Systems (AAMAS 2026)}{May 25 -- 29, 2026}
{Paphos, Cyprus}{C.~Amato, L.~Dennis, V.~Mascardi, J.~Thangarajah (eds.)}
\copyrightyear{2026}
\acmYear{2026}
\acmDOI{}
\acmPrice{}
\acmISBN{}

\acmSubmissionID{<<600>>}

\title[AAMAS-2026 Formatting Instructions]{GroupDebate: Enhancing the Efficiency of Multi-Agent Debate Using Group Discussion}

\author{Tongxuan Liu}
\affiliation{
  \institution{University of Science and Technology of China}
  \city{Beijing}
  \country{CHINA}}
\affiliation{
  \&
  \institution{JD.com}
  \city{Beijing}
  \country{CHINA}}
\email{tongxuan.ltx@mail.ustc.edu.cn}

\author{Xingyu Wang}
\affiliation{
  \institution{Institute of Automation, Chinese Academy of Sciences}
  \city{Beijing}
  \country{CHINA}}
\email{wangxingyu2024@ia.ac.cn}

\author{Weizhe Huang}
\affiliation{
  \institution{JD.com}
  \city{Beijing}
  \country{CHINA}}
\email{huangweizhe1@jd.com}

\author{Wenjiang Xu}
\affiliation{
  \institution{Institute of Automation, Chinese Academy of Sciences}
  \city{Beijing}
  \country{CHINA}}
\email{xuwenjiang2024@ia.ac.cn}

\author{Yuting Zeng}
\affiliation{
  \institution{University of Science and Technology of China}
  \city{Hefei}
  \country{CHINA}}
\email{yuting\_zeng@mail.ustc.edu.cn}

\author{Lei Jiang}
\affiliation{
  \institution{University of Science and Technology of China}
  \city{Hefei}
  \country{CHINA}}
\email{jianglei0510@mail.ustc.edu.cn}

\author{Hailong Yang}
\affiliation{
  \institution{Beihang University}
  \city{Beijing}
  \country{CHINA}}
\email{hailong.yang@buaa.edu.cn}

\author{Jing Li}
\affiliation{
  \institution{University of Science and Technology of China}
  \city{Hefei}
  \country{CHINA}}
\email{lj@ustc.edu.cn}

\begin{abstract}
    In recent years, Large Language Models (LLMs) have demonstrated remarkable capabilities across diverse NLP tasks, including complex logical reasoning, mathematical problem-solving, and multi-step decision-making. Extensive research has explored how to enhance the logical reasoning abilities such as Chain-of-Thought, Chain-of-Thought with Self-Consistency, Tree-Of-Thoughts, and multi-agent debates. In the context of multi-agent debates, significant performance improvements can be achieved with an increasing number of agents and debate rounds. However, the escalation in the number of agents and debate rounds can drastically raise the tokens cost of debates, thereby limiting the scalability of the multi-agent debate technique. To better harness the advantages of multi-agent debates in logical reasoning tasks, this paper proposes a method to significantly reduce token cost in multi-agent debates. This approach involves dividing all agents into multiple debate groups, with agents engaging in debates within their respective groups and sharing interim debate results between groups. Comparative experiments across multiple datasets have demonstrated that this method can reduce the total tokens by up to 46.9\% during debates and while potentially enhancing accuracy by as much as 21.9\%. Our method significantly enhances the performance and efficiency of interactions in the multi-agent debate.
\end{abstract}


\newcommand{\BibTeX}{\rm B\kern-.05em{\sc i\kern-.025em b}\kern-.08em\TeX}

\begin{document}


\pagestyle{fancy}
\fancyhead{}


\maketitle 


\section{Introduction}
\label{sec:introduction}

\begin{figure*}
    \centering
    \includegraphics[width=0.48\linewidth]{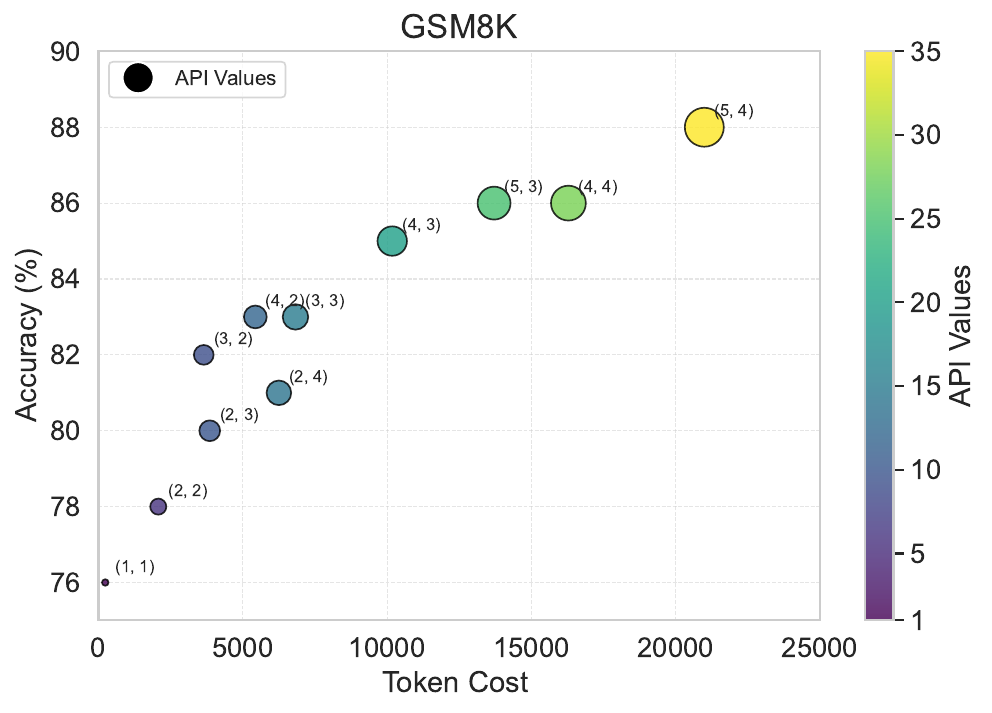}
    \hfill
    \includegraphics[width=0.48\linewidth]{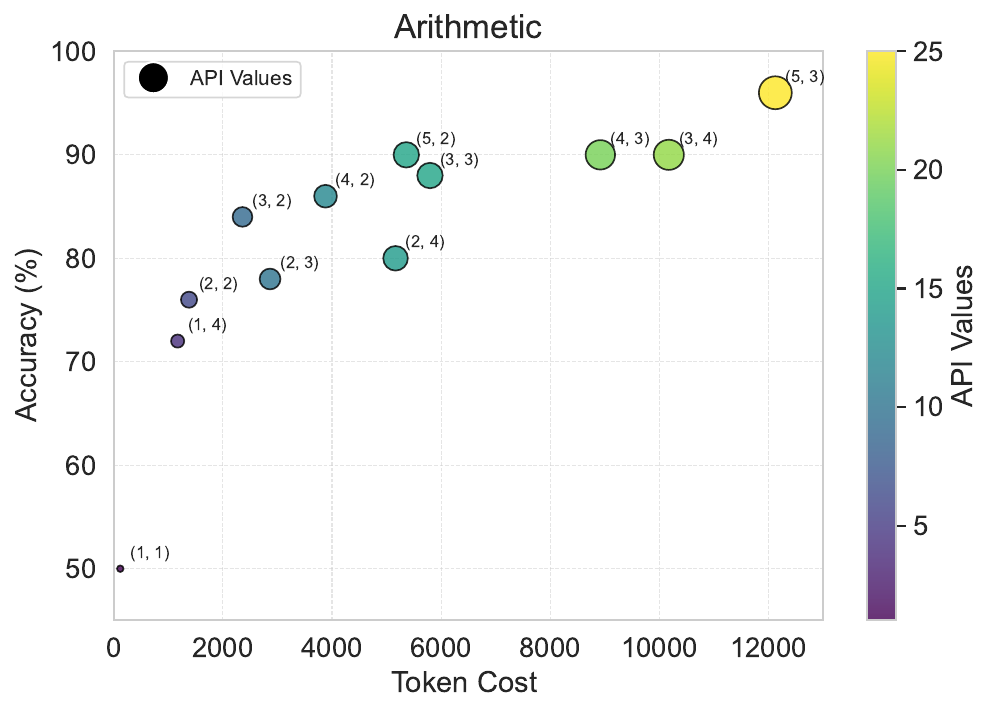}
    \caption{\textbf{Comparison of Token Cost and Accuracy Under Different Combinations of Agents and Rounds.} The numbers in parentheses corresponding to each circle represent the pair of agent number and round number. The size/color of the circle represents the number of API calls, indicating that the larger the circle, the more times the OpenAI API is called.}
    \label{fig1}
\end{figure*}

Large language Models (LLMs) such as GPT \cite{openai2024gpt4,brown2020language,bubeck2023sparks,radford2018improving,radford2019language}, LLaMa \cite{touvron2023llama,touvron2023llama2}, and PaLM \cite{anil2023palm,chowdhery2023palm} have revolutionized the field of natural language processing (NLP) by demonstrating remarkable capabilities across a wide range of tasks. These models can reach or even exceed human performance in a range of NLP tasks but their performance is still limited in complex mathematical and logical reasoning tasks \cite{liu2023evaluating}. To address these limitations, researchers have proposed developed various techniques to enhance the reasoning abilities of LLMs. Chain-of-Thought \cite{kojima2023large,wei2023chainofthought,nye2021work} encourages models to generates the reasoning process step by step. Subsequent research has introduced such as Tree-of-Thoughts \cite{yao2024tree} and Verification \cite{lightman2023lets} to enhance their ability to perform complex multi-step reasoning. Unfortunately, these single-agent methods are more likely to fall into random fabrication of facts or the generation of delusions, thus leading to erroneous outcomes \cite{bubeck2023sparks,huang2023survey,ji2023survey}. 
The multi-agent debate methods mitigate these issues by allowing different agents to express their arguments to each other and these approaches have demonstrated considerable potential and effectiveness across various types of tasks and datasets \cite{chan2023chateval,du2023improving,liang2023encouraging,sun2023corex,wang2023apollo,xiong2023examining,xu2023toward}.


However, as the number of agents and rounds increases, the token cost in multi-agent debate can escalate significantly. This issue results in monetary expenditure on tokens through LLM-based API or substantial computational overhead and power consumption, thereby severely hindering the scalability and broader application of multi-agent debate, especially in scenarios with limited computational resources \cite{guo2024large}.
As illustrated in the Figure \ref{fig1}, compared with a single LLM-based agent, employing a multi-agent debate with three agents in five rounds can potentially raise the accuracy from the initial 50\% to 98\%, but introduces 101× token cost in the Arithmetic \cite{brown2020language} task. Similarly, in the GSM8K \cite{cobbe2021training} task, five rounds of multi-agent debate involving four agents can raise the accuracy from 76\% to 88\%, but it results in 90× token cost. To address the issue of the rapidly increasing number of tokens in multi-agent debates, researchers have proposed various improved techniques. For instance, the multi-agent debate in \cite{du2023improving} summarizes the output of other agents to serve as the input for the next round. \cite{sun2023corex} proposes a "forgetfulness" mode that only the output from the previous round is stored as input for the next round. 
However, only employing a "forgetfulness" mode or summary mechanism to reduce token cost is still limited due to their theoretical complexity and the issue of exacerbated token growth. Moreover, owing to their simplistic debating modes, they struggle to fully exploit the collaborative capabilities of multi-agent debates.

In human societies, when multiple individuals engage in a debate, they usually conduct group discussion to enhance the efficiency of interaction while also preserving the diversity of viewpoints \cite{krueger2014focus}. Inspired by this, in this paper, we propose a novel method GroupDebate (GD), which leverages group discussion to further reduce token cost in multi-agent debates. Specifically, Our method divides all participating agents into several debate groups, with each group conducting internal debates. Following the debates, the results are summarized and placed into a shared pool. After that, each group of agents retrieves the debate summaries of all groups from the pool, which serve as the input for the agents in the next round. Upon the conclusion of the debate, all agents reach a consensus or the final outcome is determined by majority vote. Furthermore, we conduct a theoretical analysis of the total token cost of the GroupDebate, thereby affirming the effectiveness of the method. In our experiments, we evaluate the effectiveness of GroupDebate in comparison to existing multi-agent debate methods and observe up to 34.8\%/45.2\%/46.9\%/39.3\%/30.6\% reduction in token cost in the Arithmetic/GSM8K/MMLU/MATH/GPQA dataset, as well as up to 12\%/21.9\% improvement in accuracy in the MMLU/GPQA dataset. 

Our main contributions are as follows: 

\begin{itemize}
    \item[1.] We propose an innovative multi-agent debate strategy GroupDebate based on group discussion which can improve the efficiency and performance of multi-agent debates.
    \item[2.] We conduct a theoretical analysis of token cost based on our method, demonstrating its efficiency and effectiveness.
    \item[3.] Extensive experiments across five logical reasoning and mathematical datasets show that our method can not only significantly reduce token cost but also potentially enhance accuracy, validating the efficiency and superiority of our method.
\end{itemize}
\section{Preliminaries}
\label{sec:preliminaries}

\begin{figure*}[t]
    \centering
    \includegraphics[width=1\linewidth]{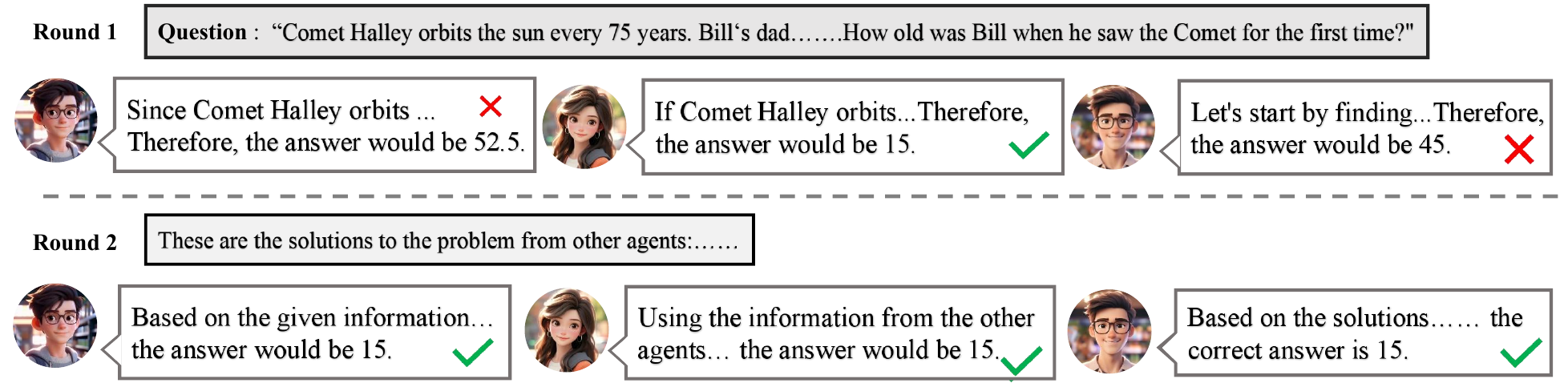}
    \caption{\textbf{An Example of Multi-agent Debate Among Three Agents with Two Rounds.} }
    \label{fig2}
\end{figure*}

\subsection{Multi-agent Debate}

In the context of multi-agent debates (MAD), by integrating multiple LLMs (each treated as an individual agent) and using various collaboration strategies, agents can propose viewpoints, review, and respond to the results of other agents in multiple rounds of debates \cite{chan2023chateval,sun2023corex,terekhov2023second}. The process of MAD can be summarized as follows:

(i) At the beginning, each agent is provided with a question and generates an individual response;

(ii) These responses then form the new input context for each agent, and the agents generate new responses; 

(iii) This debate procedure is repeated over multiple rounds and the final answer is obtained through majority voting.

Throughout multi-agent debate procedure, all agents can consistently improve their own responses based on the responses of other agents. In order to reduce input context length, \cite{du2023improving} proposes that after collecting the responses from other agents, the responses should first be summarized and then used as the new input context for each agent. 
Figure \ref{fig2} shows an example of two-round debates among three agents. In the first round, each agent independently responds to the input and their outputs are collected and summarized. In the second round, each agent's input includes summaries from the previous round, which are combined with a prompt to guide the output. 
Ultimately, all agents reach a consensus conclusion.

\subsection{Token Cost in Multi-agent Debate}

\begin{figure}
    \centering
    \includegraphics[width=0.8\linewidth]{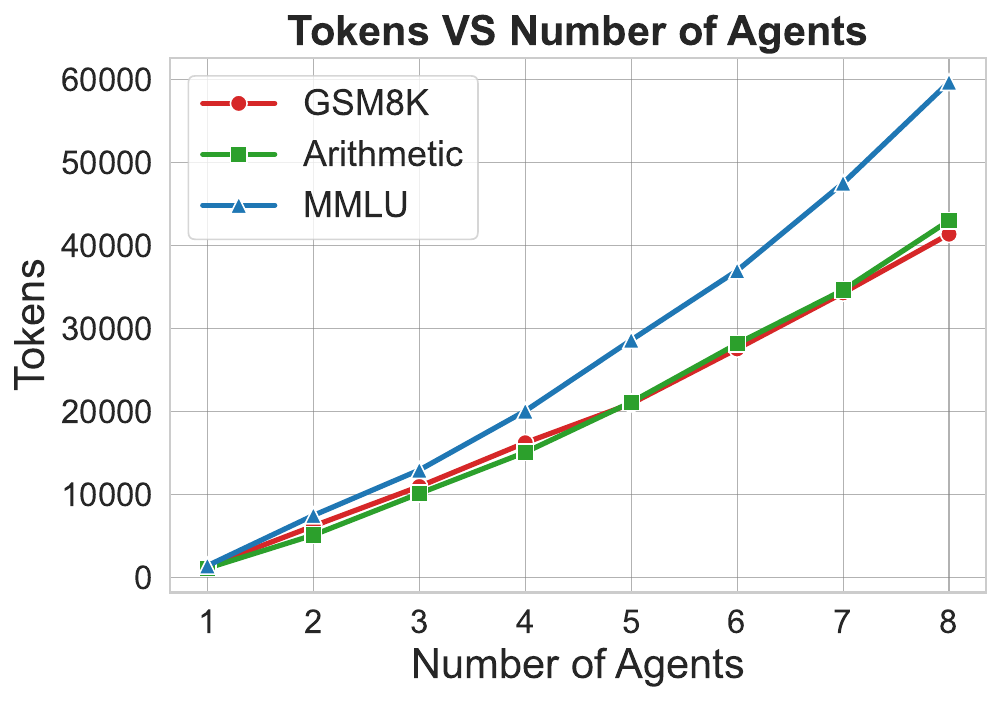}
    \includegraphics[width=0.8\linewidth]{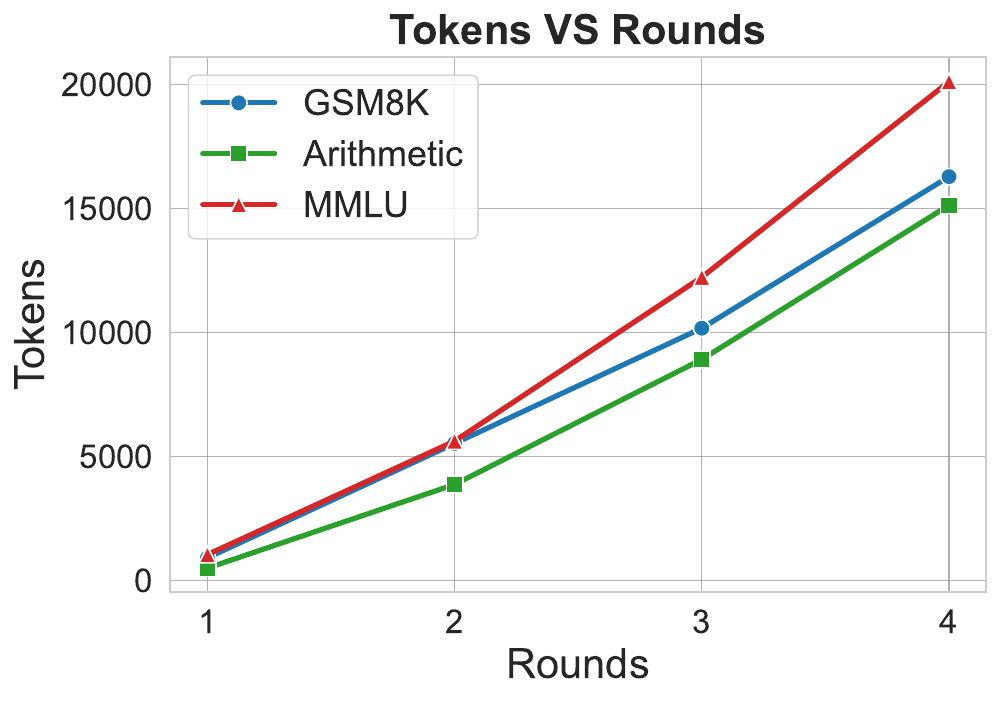}
    \caption{\textbf{Token Cost Under Different Numbers of Agents and Rounds.} The upper figure illustrates the token cost with variations in agents under the premise of 4 rounds. The lower figure illustrates the token cost with changes in rounds under the condition of 4 agents.}
    \label{fig3}
\end{figure}

In the Figure \ref{fig1}, we can observe that although an increase in the number of agents and rounds can significantly enhance accuracy, the sharply increasing token cost is still a serious challenge in multi-agent debate. We further analyze this token cost issue based on the Simultaneous-Talk interaction strategy \cite{chan2023chateval}, where each agent synchronizes their results with other agents in each round of the debate. From Figure \ref{fig3}, it can be observed that under 4 rounds, as the number of agents increases from 1 to 8, the token cost in GSM8K/Arithmetic/MMLU has respectively grown by 36×/44×/49×. Similarly, under 4 agents, as the number of rounds increases from 1 to 4, the token cost in GSM8K/Arithmetic//MMLU has respectively increased by 17×/29×/19×. 


\section{Methodology}
\label{sec:methodology}

\begin{figure*}[t]
    \centering
    \includegraphics[width=1\linewidth]{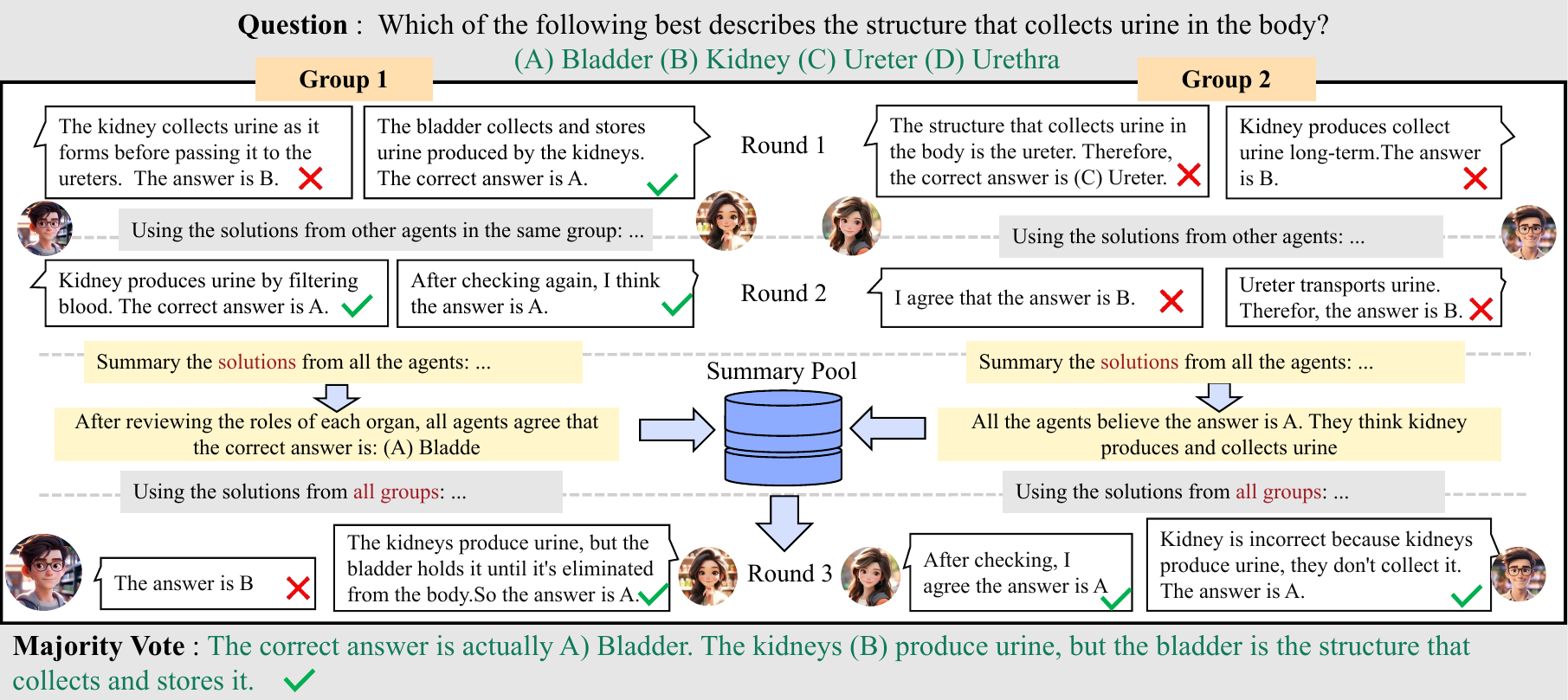}
    \caption{\textbf{An Example of GroupDebate.} 4 agents are divided into 2 groups and the GroupDebate process comprises two stages, with each stage involving two rounds of intra-group debate.}
    \label{fig4}
    \vspace{-0.5cm}
\end{figure*}

In this section, we first introduce the overall framework of our GroupDebate. Subsequently, we provide mathematical analysis of the token cost for both MAD and our GroupDebate. Formally, assume there are $M$ LLM-based agents, denoted as $A=\{A_i | i=1,2,\cdots,M\}$, participating in a multi-round debate, with the total number of debate rounds denoted as $T$. In each round $t$ $(t=1,2,\ldots,T)$, the output of each agent $A_i$ is represented as $Output_i^{t}$. These outputs are dynamically refined through structured inter-agent interactions, where agents critique, verify, and build upon reasoning from others. The tokens of the initial question prompt are denoted as $Q$, serves as the foundational query propagated through the debate process. These notations will be used throughout this paper.

\subsection{GroupDebate} 
\label{multi-agent-group-debate}

The GroupDebate framework orchestrates collaborative reasoning among 
M LLM-based agents, denoted as $A=\{A_i | i=1,2,\cdots,M\}$, which can be randomly divided into $N$ groups $G=\{G_j|j=1,2,\cdots,N\}$, with average K agents in each group. 
The GroupDebate splits the total debate rounds into $S$ stages, with each stage encompassing $R$ rounds. Thus, the total number of rounds $T$ can be calculated as $T=S\times R$. This hierarchical design enables efficient exploration of solution spaces through alternating phases of localized refinement (intra-group) and global synthesis (inter-group).
For the $s$-th stage's $r$-th round, GroupDebate selects one of the following processes:
\begin{enumerate}
\item[(1)] \textbf{Initial Thinking.} If $s=1$ and $r=1$ (i.e., the first stage's first round), we input the initial question prompt $Q$ to each agent.
\item[(2)] \textbf{Intra-group Debate.} If $r>1$, we utilize the outputs from other agents within the same group as the input for each agent.
\item[(3)] \textbf{Inter-group Debate.} If $s>1$ and $r=1$, we merge the outputs from the last round of each group into a summary and input the summaries from other groups to each agent.
\end{enumerate}

Meanwhile, inspired by \cite{sun2023corex}, we summarize the responses from other groups and restrict each agent to receive the latest summary from the previous stage in the inter-group debate.
After the $S$-th stage's $R$-th round, all agents vote, and the ultimate result is determined by the majority selection. The detailed GroupDebate process can be found in Appendix \ref{appendix:gd-al}. The Figure \ref{fig4} illustrates an example of GroupDebate consisting of two stages and two groups. In the first stage, two agents in each group receive the initial question and exchange ideas within the group. In the second stage, agents share the summaries of their respective groups between groups and then discuss within their own groups again.


\subsection{Token Cost Analysis} \label{token-consumption-analysis}

\paragraph{Token Cost in Multi-agent Debate.} 

We implement the summary mechanism in MAD following \cite{du2023improving}, where the output of other agents is summarized and used as input for each agent in the next round. The summary for agent $A_i$ in round $t$ is denoted as $Summary_i^t$. 
Then token cost $Token^t$ in each round $t$ can be computed as follows:

\begin{equation}
\begin{split}
\left\{
\begin{aligned}
& \sum_{i=1}^M (Q+Output_i^t),   &&{t=1} \\
& Token^{t-1}+\sum_{i=1}^{M}(S_{i}^{t-1}+  Output_{i}^{t}),   &&{t>1}
\end{aligned}
\right.
\end{split}
\end{equation}
Finally, the total token cost in MAD is 
\begin{align}
Token^{GD} = \mathcal{O}\left( MTQ+(M^2T+MT^2)C \right)
\end{align}
where $C$ represents the upper bound on the token number for each agent's response and the generated summary. More mathematical details are illustrated in Appendix \ref{appendix:token-cost-mad}.

\paragraph{Token Cost in GroupDebate.} In GroupDebate, we summarize the outputs from other groups at the end of each stage. Here, we define the summary of group $G_j$ at the end of stage $s$ as $Summary_j^s$. 

Then token cost $Token^t_s$ in round $t$ at stage $s$ is:

Then the token cost in the intra-group debate is:
\begin{equation}
    \sum_{j=1}^N \sum_{i\in G_j} (Q+Output_i^{t}+ \sum_{i'\in G_j} Output_{i'}^{t-1}),
\end{equation}
where $(s-1)R+1<t<=\min(sR,T)$. The token cost in the inter-group debate is:
\begin{multline}
    \sum_{i=1}^{M} (Q + Output_i^{t-1} + \sum_{j=1}^{N} Summary_j^{s-1} + Output_i^{t}),
\end{multline}
where $t=(s-1)R+1$.
Finally, the total token cost of GroupDebate is $Token^{GD} = \mathcal{O}\left(MTQ+(\frac{M^2T}{N}+MSN)C\right)$, where $C$ represents the upper bound on the token number for each agent's response and the generated summary. More calculation details are shown in Appendix \ref{appendix:token-cost-gd}.

\begin{figure*}[t]
    \centering
    \includegraphics[width=1\linewidth]{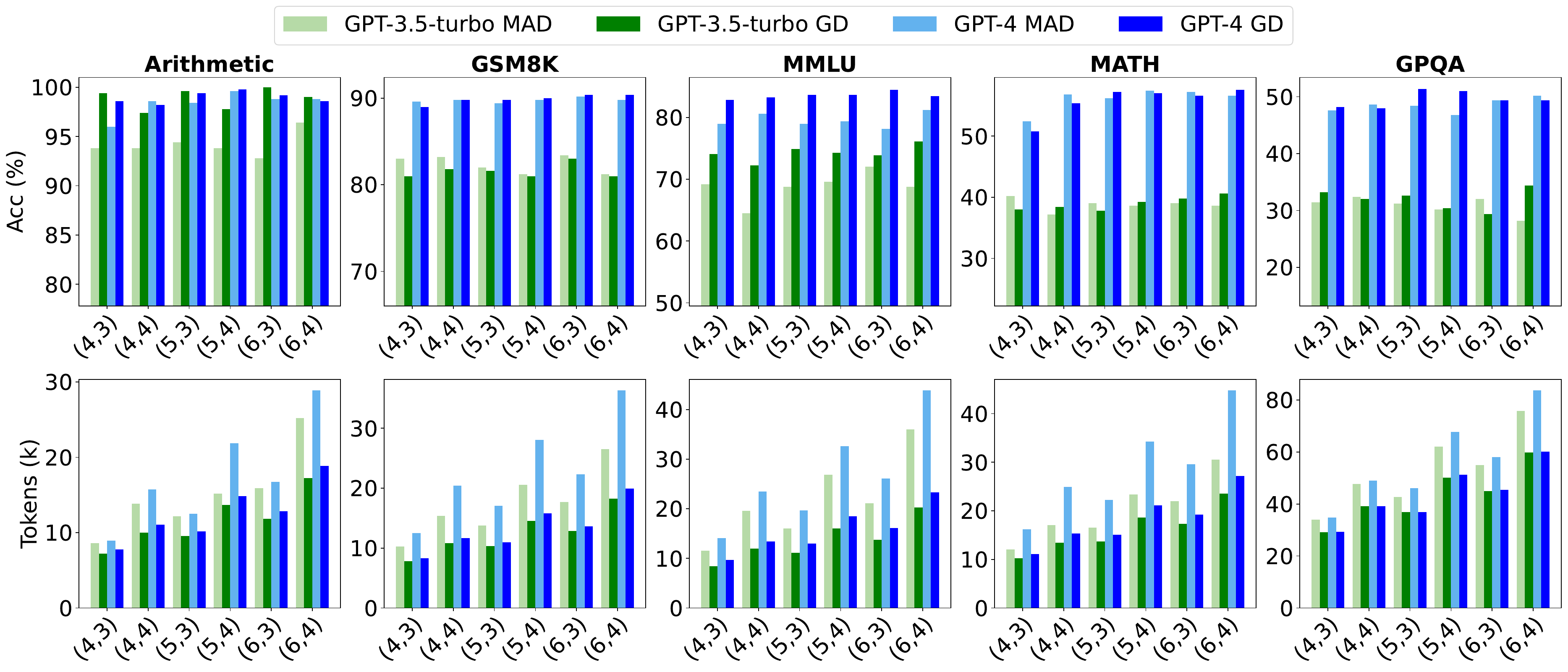}
    \caption{\textbf{Comparison of Token Cost and Accuracy Between GD and MAD under Different Agents and Rounds.} The notation (5,4) signifies 5 agents with 4 rounds. The results are average.}
    \label{fig5}
\end{figure*}

\paragraph{Discussion.} From the perspective of overall token cost complexity, GD and MAD exhibit the same level of complexity regarding the input token cost of the question prompt $Q$, indicating that the question prompt has an equal impact on both methods. In our GroupDebate, given fixed values for $T$ and $M$, the number of groups $N$ and the total number of stages $S$ can be dynamically adjusted. When we set $N \rightarrow \mathcal{O}\left(\sqrt{\frac{MT}{S}}\right)$, theoretically, we can obtain $Token^{GD} \rightarrow \mathcal{O}\left(MTQ+\sqrt{M^3TS}C\right)$. This complexity is significantly lower than that of MAD. If we consider setting $S$ to a small positive integer, treating it as a constant, then $Token^{GD}$ can further approach $\mathcal{O}\left(MTQ+\sqrt{M^3T}C\right)$. Moreover, $N$ and $S$ also influence the diversity in multi-agent debate, affecting the accuracy of the debate results, which will be further studied in Section \ref{hyperparameter-study}.

\section{Experiments}
\label{sec:experiments}

\subsection{Experimental Setup} 


\begin{table*}[t]
\centering
\setlength\tabcolsep{1.5 pt}
\resizebox{\textwidth}{!}{
\begin{tabular}{ccccccccccc} 
\toprule
\multirow{2}{*}{Methods} & \multicolumn{2}{c}{GPQA} & \multicolumn{2}{c}{GSM8k} & \multicolumn{2}{c}{Arithmetic} & \multicolumn{2}{c}{MMLU} & \multicolumn{2}{c}{Math}  \\ 
\cmidrule{2-11}
& ACC(\%)$\uparrow$  & Tokens($k$)$\downarrow$   & ACC(\%)$\uparrow$   & Tokens($k$)$\downarrow$   & ACC(\%)$\uparrow$   & Tokens($k$)$\downarrow$       & ACC(\%)$\uparrow$    & Tokens($k$)$\downarrow$      & ACC(\%)$\uparrow$   & Tokens($k$)$\downarrow$    \\ 
\midrule
\rowcolor{Gray}
\multicolumn{11}{c}{\texttt{GPT-3.5-turbo-0125}} 
\\
\midrule
CoT                      & 31.2\scriptsize{$\pm 0.03$}  & 2.0\scriptsize{$\pm 0.02$}             & 76.8\scriptsize{$\pm 0.02$} & 0.25\scriptsize{$\pm 0.00$}         & 82.2\scriptsize{$\pm 0.04$} & 0.16\scriptsize{$\pm 0.02$}                & 70.2\scriptsize{$\pm 0.02$}  & 0.24\scriptsize{$\pm 0.00$}         & 35.2\scriptsize{$\pm 0.02$} & 0.37\scriptsize{$\pm 0.01$}           \\
CoT-SC(40)               & 31.2\scriptsize{$\pm 0.02$} & 80.5\scriptsize{$\pm 0.26$}          & \underline{\textbf{83.6}}\scriptsize{$\pm 0.01$} & \underline{10.0}\scriptsize{$\pm 0.02$}        & 95.0\scriptsize{$\pm 0.01$}  & 6.3\scriptsize{$\pm 0.11$}                 & \underline{\textbf{75.1}}\scriptsize{$\pm 0.02$}  & \underline{\textbf{9.6}}\scriptsize{$\pm 0.03$}         & \textbf{48.2}\scriptsize{$\pm 0.01$} & 14.6\scriptsize{$\pm 0.15$}           \\
MAD(5,3)                 & 31.2\scriptsize{$\pm 0.05$} & 42.7\scriptsize{$\pm 0.79$}          & 82.0\scriptsize{$\pm 0.01$}  & 13.7\scriptsize{$\pm 0.15$}        & 94.4\scriptsize{$\pm 0.01$} & 12.1\scriptsize{$\pm 0.27$}                & 68.8\scriptsize{$\pm 0.01$} & 16.0\scriptsize{$\pm 0.10$}         & 39.0\scriptsize{$\pm 0.02$}  & 16.5\scriptsize{$\pm 0.23$}           \\ 
GD(5,3)                  & \underline{\textbf{32.6}}\scriptsize{$\pm 0.05$} & \underline{36.9}\scriptsize{$\pm 0.46$}          & 81.6\scriptsize{$\pm 0.01$} & 10.3\scriptsize{$\pm 0.05$}         & \textbf{99.6}\scriptsize{$\pm 0.00$} & 9.6\scriptsize{$\pm 0.06$}                & 74.9\scriptsize{$\pm 0.03$} & 11.1\scriptsize{$\pm 0.07$}         & 37.8\scriptsize{$\pm 0.00$} & 13.7\scriptsize{$\pm 0.10$}           \\ 
\midrule
\rowcolor{Gray}
\multicolumn{11}{c}{\texttt{GPT-4-0613}} 
\\
\midrule
CoT                      & 50.2\scriptsize{$\pm 0.02$} & 2.0\scriptsize{$\pm 0.01$}           & 88.8\scriptsize{$\pm 0.01$} & 0.25\scriptsize{$\pm 0.00$}         & 94.4\scriptsize{$\pm 0.02$} & 0.20\scriptsize{$\pm 0.00$}                & 75.1\scriptsize{$\pm 0.02$}  & 0.28\scriptsize{$\pm 0.00$}         & 46.4\scriptsize{$\pm 0.05$} & 0.39\scriptsize{$\pm 0.01$}           \\
CoT-SC(40)               & 47.6\scriptsize{$\pm 0.02$} & 80.8\scriptsize{$\pm 0.18$}          & \underline{\textbf{91.0}}\scriptsize{$\pm 0.00$}  & \underline{10.0}\scriptsize{$\pm 0.01$}         & \underline{\textbf{100.0}}\scriptsize{$\pm 0.00$}     & \underline{8.3}\scriptsize{$\pm 0.03$}                & 78.8\scriptsize{$\pm 0.02$} & 5.7\scriptsize{$\pm 0.08$}         & \textbf{62.8}\scriptsize{$\pm 0.00$} & 15.3\scriptsize{$\pm 0.12$}           \\
MAD(5,3)                 & 48.4\scriptsize{$\pm 0.02$} & 46.1\scriptsize{$\pm 2.07$}          & 89.4\scriptsize{$\pm 0.00$} & 17.1\scriptsize{$\pm 0.06$}        & 98.4\scriptsize{$\pm 0.01$} & 12.5\scriptsize{$\pm 0.01$}                & 79.0\scriptsize{$\pm 0.02$} & 20.0\scriptsize{$\pm 0.11$}         & 56.2\scriptsize{$\pm 0.02$} & 22.3\scriptsize{$\pm 0.24$}           \\ 
GD(5,3)                  & \underline{\textbf{51.4}}\scriptsize{$\pm 0.03$} & \underline{36.9}\scriptsize{$\pm 1.11$}           & 89.8\scriptsize{$\pm 0.02$} & 11.0\scriptsize{$\pm 0.01$}         & 99.4\scriptsize{$\pm 0.01$} & 10.1\scriptsize{$\pm 0.33$}                & \textbf{83.7}\scriptsize{$\pm 0.01$} & 13.0\scriptsize{$\pm 0.12$}         & 57.2\scriptsize{$\pm 0.03$} & 15.1\scriptsize $\pm 0.19$ \\
\midrule
\rowcolor{Gray}
\multicolumn{11}{c}{\texttt{DeepSeek-R1-Distill-Qwen-32B}} 
\\
\midrule
CoT                      & 30.3\scriptsize{$\pm 0.02$} & 3.27\scriptsize{$\pm 0.57$}           & 90.7\scriptsize{$\pm 0.02$} & 0.58\scriptsize{$\pm 0.00$}         & 99.0\scriptsize{$\pm 0.01$} & 0.55\scriptsize{$\pm 0.00$}                & 79.9\scriptsize{$\pm 0.01$}  & 0.90\scriptsize{$\pm 0.02$}         & 56.3\scriptsize{$\pm 0.03$} & 1.62\scriptsize{$\pm 0.02$}           \\
CoT-SC(40)               & 57.7 \scriptsize{$\pm 0.02$} & 143.1 \scriptsize{$\pm 1.51$}          & 92.7\scriptsize{$\pm 0.00$}  & \underline{23.6}\scriptsize{$\pm 0.12$}         & \underline{\textbf{100}}\scriptsize{$\pm 0.00$}     & \underline{22.4}\scriptsize{$\pm 0.04$}                & 83.3\scriptsize{$\pm 0.00$} & 36.5\scriptsize{$\pm 0.01$}         & 71.7\scriptsize{$\pm 0.03$} & 64.5\scriptsize{$\pm 0.09$}           \\
MAD(5,3)                 & \underline{\textbf{60.7}}\scriptsize{$\pm 0.02$} & 125.5\scriptsize{$\pm 2.12$}          & 92.7\scriptsize{$\pm 0.01$} & 57.9\scriptsize{$\pm 0.28$}        & 100\scriptsize{$\pm 0.00$} & 65.8\scriptsize{$\pm 0.50$}                & 84.0\scriptsize{$\pm 0.02$} & 66.5\scriptsize{$\pm 0.52$}         & 87.0\scriptsize{$\pm 0.02$} & 104.8\scriptsize{$\pm 0.24$}           \\ 
GD(5,3)                  & 59.0\scriptsize{$\pm 0.01$} & \underline{78.4}\scriptsize{$\pm 3.23$}           & \underline{\textbf{93.3}}\scriptsize{$\pm 0.01$} & 31.9\scriptsize{$\pm 0.48$}         & 100\scriptsize{$\pm 0.00$} & 33.5\scriptsize{$\pm 0.33$}                & \underline{\textbf{85.4}}\scriptsize{$\pm 0.01$} & \underline{31.9}\scriptsize{$\pm 0.21$}         & \underline{\textbf{89.0}}\scriptsize{$\pm 0.01$} & \underline{58.1}\scriptsize{$\pm 0.58$} \\
\rowcolor{Gray}
\multicolumn{11}{c}{\texttt{DeepSeek-R1}} 
\\
\midrule
CoT                      & 39.3\scriptsize{$\pm 0.00$} & 3.80\scriptsize{$\pm 0.00$}           & 93.2\scriptsize{$\pm 0.01$} & 0.94\scriptsize{$\pm 0.00$}         & 95.2\scriptsize{$\pm 0.00$} & 0.73\scriptsize{$\pm 0.01$}                & 84.0\scriptsize{$\pm 0.01$}  & 1.04\scriptsize{$\pm 0.01$}         & 69.2\scriptsize{$\pm 0.01$} & 1.64\scriptsize{$\pm 0.02$}           \\
CoT-SC(40)               & 61.7 \scriptsize{$\pm 0.02$} & 140.2 \scriptsize{$\pm 1.21$}          & 94.2\scriptsize{$\pm 0.00$}  & 37.6\scriptsize{$\pm 0.21$}         & \underline{\textbf{100}}\scriptsize{$\pm 0.00$}     & 22.4\scriptsize{$\pm 0.04$}                & 88.2\scriptsize{$\pm 0.01$} & 42.6\scriptsize{$\pm 0.01$}         & 78.2\scriptsize{$\pm 0.01$} & 65.4\scriptsize{$\pm 0.04$}           \\
MAD(5,3)                 & 62.0\scriptsize{$\pm 0.01$} & 148.2\scriptsize{$\pm 1.13$}          & \underline{\textbf{95.2}}\scriptsize{$\pm 0.02$} & 61.7\scriptsize{$\pm 0.23$}        & 99.0\scriptsize{$\pm 0.01$} & 55.3\scriptsize{$\pm 0.30$}                & 86.7\scriptsize{$\pm 0.02$} & 81.4\scriptsize{$\pm 0.12$}         & 90.2\scriptsize{$\pm 0.02$} & 84.5\scriptsize{$\pm 0.12$}           \\ 
GD(5,3)                  & \underline{\textbf{63.0}}\scriptsize{$\pm 0.01$} & \underline{72.9}\scriptsize{$\pm 0.02$}           & 94.4\scriptsize{$\pm 0.01$} & \underline{26.6}\scriptsize{$\pm 0.28$}         & 99.4\scriptsize{$\pm 0.00$} & \underline{12.9}\scriptsize{$\pm 0.13$}                & \underline{\textbf{90.2}}\scriptsize{$\pm 0.02$} & \underline{34.4}\scriptsize{$\pm 0.01$}         & \underline{\textbf{90.2}}\scriptsize{$\pm 0.01$} & \underline{47.9}\scriptsize{$\pm 0.08$} \\
\rowcolor{Gray}
\multicolumn{11}{c}{\texttt{DeepSeek-V3}} 
\\
\midrule
CoT                      & 50.7\scriptsize{$\pm 0.00$} & 2.15\scriptsize{$\pm 1.3$}           & 95.2\scriptsize{$\pm 0.02$} & 0.36\scriptsize{$\pm 0.01$}         & 100\scriptsize{$\pm 0.01$} & 0.25\scriptsize{$\pm 0.00$}                & 86.3\scriptsize{$\pm 0.01$}  & 0.41\scriptsize{$\pm 0.02$}         & 76.4\scriptsize{$\pm 0.03$} & 0.84\scriptsize{$\pm 0.02$}           \\
CoT-SC(40)               & 50.0 \scriptsize{$\pm 0.04$} & 87.1 \scriptsize{$\pm 6.4$}          & 94.2\scriptsize{$\pm 0.02$}  & 37.6\scriptsize{$\pm 0.42$}         & \underline{\textbf{100}}\scriptsize{$\pm 0.00$}     & \underline{9.88}\scriptsize{$\pm 0.04$}                & \underline{\textbf{88.4}}\scriptsize{$\pm 0.01$} & \underline{16.4}\scriptsize{$\pm 0.12$}         & 84.4\scriptsize{$\pm 0.03$} & 33.2\scriptsize{$\pm 0.39$}           \\
MAD(5,3)                 & 54.0\scriptsize{$\pm 0.02$} & 74.8\scriptsize{$\pm 2.37$}          & \underline{\textbf{95.6}} \scriptsize{$\pm 0.01$} & 33.0\scriptsize{$\pm 0.26$}        & 100\scriptsize{$\pm 0.00$} & 29.1\scriptsize{$\pm 0.20$}                & 86.7\scriptsize{$\pm 0.01$} & 42.0\scriptsize{$\pm 0.22$}         & 85.2\scriptsize{$\pm 0.12$} & 33.0\scriptsize{$\pm 0.14$}           \\ 
GD(5,3)                  & \underline{\textbf{54.0}}\scriptsize{$\pm 0.02$} & \underline{45.6}\scriptsize{$\pm 1.23$}           & 94.4\scriptsize{$\pm 0.01$} & \underline{13.8}\scriptsize{$\pm 0.31$}         & 100\scriptsize{$\pm 0.00$} & 10.4\scriptsize{$\pm 0.13$}                & 87.3\scriptsize{$\pm 0.01$} & 17.3\scriptsize{$\pm 0.04$}         & \underline{\textbf{86.0}}\scriptsize{$\pm 0.01$} & \underline{24.4}\scriptsize{$\pm 0.12$} \\
\bottomrule
\end{tabular}
}
\caption{\textbf{Comparison of Token Cost and Accuracy Between GD and Other Methods.} The results of highest accuracy are \textbf{bold} and the results of both highest accuracy and lowest token cost except from CoT are \underline{underlined}}
\label{comp_with_single}
\end{table*}

\paragraph{Tasks and Metrics.} To demonstrate the accuracy and effectiveness of different methods, we adopt total token cost and accuracy (ACC) as evaluation metrics. Additionally, we select five representative tasks related to logical reasoning and mathematical tasks to evaluate our methods, namely Arithmetic \cite{brown2020language}, GSM8K \cite{cobbe2021training}, MMLU \cite{hendrycks2020measuring}, MATH \cite{hendrycks2021measuring} and GPQA \cite{rein2023gpqa}.

\paragraph{Baselines.} To evaluate the performance of GroupDebate (GD), We conduct a comparison of the efficiency and accuracy between GD and the following methods: (1) Chain-of-Thought (CoT) \cite{wei2023chainofthought}, which employs a sequential reasoning process to generate solutions; 
(2) Self-Consistency with Chain-of-Thought (CoT-SC) \cite{wang2023selfconsistency}, an enhanced version of CoT that aggregates multiple reasoning paths to improve robustness, where CoT-SC(40) specifically denotes the use of 40 reasoning paths; (3) multi-agent debate (MAD) \cite{liang2023encouraging}, a collaborative approach where multiple agents engage in iterative discussions to refine their solutions. For MAD, we explore various configurations by varying the number of agents and debate rounds. For example, both GD(5,3) and MAD(5,3) indicate configurations utilizing 5 agents and 3 rounds, allowing for a direct comparison of their performance under identical experimental conditions.

\paragraph{Implementation Details.} We set the number of rounds of intra-group debate to 2 in GD. Additionally, we only retain output from the last round or summary generated from the last stage. Our experiments are conducted using the following models: \texttt{GPT-3.5-turbo} \texttt{-0301}, \texttt{GPT-4-0613}, \texttt{DeepSeek-R1-Distill-Qwen-32B}, \texttt{DeepSeek-R1} and \texttt{DeepSeek-V3}. In order to prevent the input prompt token exceeding the context limit, the MAD defaults to using the summary \cite{du2023improving}. For all baselines and GD, we conduct five independent experiments separately and calculate the average. We evaluate these methods in a zero-shot setting, and the details about prompts are illustrated in Appendix \ref{appendix:prompts}.

\subsection{Main Results} \label{sec:exp-main-res}

In this section, we present a detailed comparison of GroupDebate (GD) with multi-agent debate (MAD) and other single-agent methods, including Chain-of-Thought (CoT) and Self-Consistency with Chain-of-Thought (CoT-SC(40)).  Notably, in the MATH dataset, MAD fails to produce results in both the (6,3) and (6,4) configurations due to the input prompt tokens exceeding the context limit of GPT-3.5. The key observations from our experiments are as follows: 
\paragraph{Comparison Between GD and MAD.} First, as illustrated in Figure \ref{fig5}, GD consistently reduces token cost across different models under different agent and round settings, especially achieving up to 34.8\%/45.2\%/46.9\%/39.3\%/30.6\% reduction in token cost in the Arithmetic/GSM8K/MMLU/MATH/GPQA datasets. This demonstrates that our method can effectively reduce token cost in multi-agent debate while being theoretically grounded. Second, GD also improves accuracy in most settings, achieving up to 4.6\%/4.9\%/1.2\% improvement in accuracy in certain settings of the Arithmetic/MMLU/GPQA dataset, which suggests GD can potentially enhance accuracy in multi-agent debate while reducing much token cost.

\paragraph{Comparison Between GD and Other Methods.} As shown in Table \ref{comp_with_single}, GD(5,3) and MAD(5,3) can significantly outperforms the standard single-agent method CoT, demonstrating the superiority of multi-agent debate in terms of accuracy. Moreover, multi-agent methods generally incur higher token cost compared to single-agent methods, indicating a significant challenge in reducing token cost among them. Our GD method can achieve the lowest token cost while ensuring high accuracy among all methods except CoT. Besides, we observe that CoT-SC(40) can achieve comparable performance of both accuracy and token cost with GD in average.

\subsection{In-Depth Analysis} \label{hyperparameter-study}

\begin{table}
\centering
\begin{tabular}{ccc} 
\toprule
 Strategy    & ACC(\%)$\uparrow$    & Tokens($k$)$\downarrow$  \\ 
\midrule
MAD(6,4)   & 68.8\scriptsize{$\pm 0.02$} & 36.0\scriptsize{$\pm 0.23$}  \\
\midrule
3 + 3   & \textbf{76.1}\scriptsize{$\pm 0.03$} & 20.3\scriptsize{$\pm 0.17$}  \\
2 + 2 + 2 & 74.7\scriptsize{$\pm 0.02$} & \textbf{18.6}\scriptsize{$\pm 0.09$}  \\
4 + 2   & 75.3\scriptsize{$\pm 0.01$}  & 21.3\scriptsize{$\pm 0.05$}  \\
\bottomrule
\end{tabular}
\caption{\textbf{Comparison of different group strategies within GD(6,4) on the MMLU dataset.} The notation 3+3 denotes two groups, each consisting of three agents. The best results are \textbf{bold}.}
\label{group}
\end{table}

\paragraph{Group Strategy.} In order to investigate the impact of different group strategy on accuracy and token cost, we conduct comparative experiments using GD(6,4) and \texttt{GPT-3.5-turbo-0125} on the MMLU dataset. As illustrated in Table \ref{group}, different group strategies can consistently enhance accuracy performance and reduce token cost compared to not grouping, demonstrating the effectiveness of group discussion. Moreover, 3+3 achieves the best accuracy in our experiment, indicating that more groups do not always mean the better accuracy performance. We leave the exploration of optimal group strategy parameters for achieving the best accuracy to future work.

\begin{figure}[t]
    \centering
    \includegraphics[width=0.8\columnwidth]{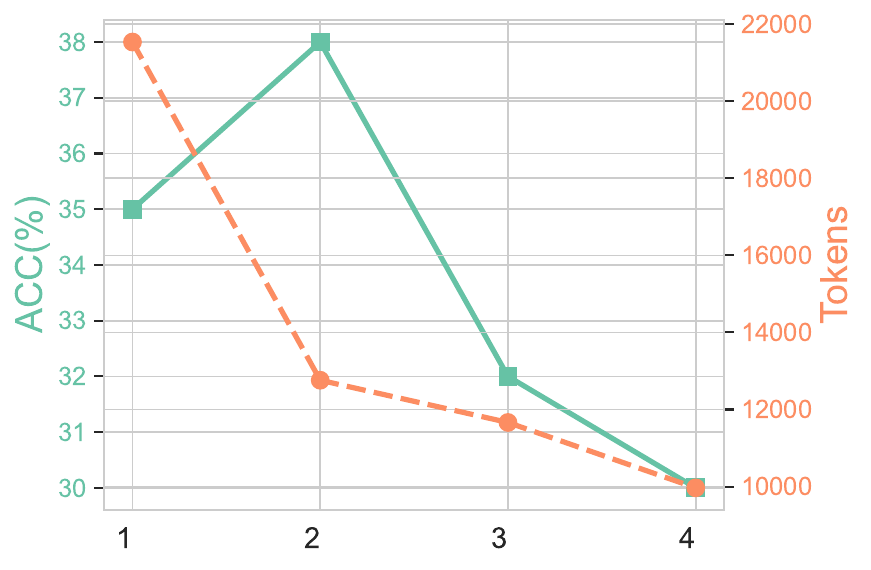}
    \caption{\textbf{Different Intra-group Debate Rounds.} The variations in accuracy are brought about by different intra-group rounds $R$.}
    \label{fig8}
\end{figure}

\paragraph{Intra-group Debate Rounds.} To explore the impact of the number of intra-group debate rounds, we conduct analysis under the condition of 4 agents and 4 rounds with varying numbers of intra-group debate rounds. As shown in Figure \ref{fig8}, best accuracy can be achieved when the number of intra-group debate rounds $R$ is 2. This suggests that brief intra-group discussion can achieve better accuracy. Moreover, as $R$ increases, the number of stages $S$ decreases, resulting in lower token cost, which aligns with our derived complexity formula.

\subsection{Scaling Study}
\begin{figure}[t]
    \centering
    \includegraphics[width=1\columnwidth]{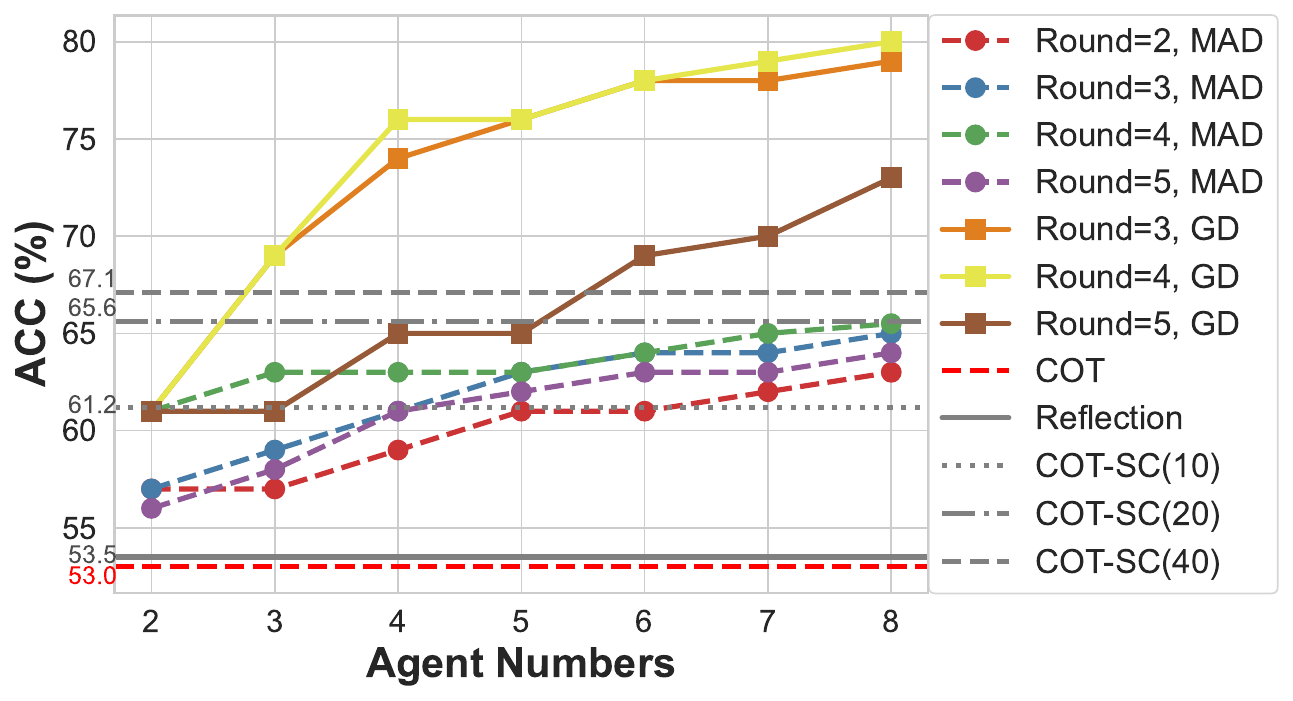}
    \caption{\textbf{Scaling Study of Agents and Rounds.} }
    \label{fig10}
\end{figure}

\begin{figure*}
    \centering
    \includegraphics[width=1\linewidth]{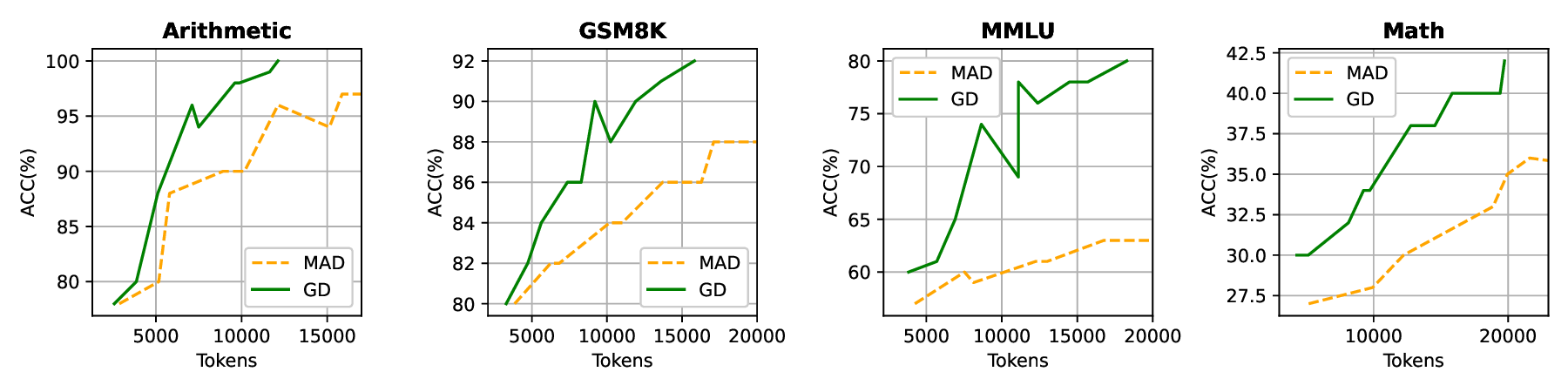}
    \caption{\textbf{Scaling Study of Token Cost.} }
    \label{fig_scaling_token}
\end{figure*}

\paragraph{Agent and Round Scaling.} In order to explore the influence of rounds and agents on accuracy under MAD and GD, we evaluate the changing trends of accuracy for MAD and GD under various rounds and agents. As shown in Figure \ref{fig10}, with the increase in rounds, there is a significant growth in accuracy, but when rounds exceeds 4, a decrease in accuracy is observed across different numbers of agents. This reflects the phenomenon that limited increase in rounds can enhance accuracy, but excessive debate rounds can lead to accuracy degradation. As the number of agents increases, there is a significant growth in accuracy, indicating that an increase in agents can notably enhance the accuracy for both MAD and GD. Concurrently, it should be noted that the rate of improvement in accuracy tends to gradually decelerate as the number of agents continues to rise. The experimental results indicate the importance of controlling the appropriate number of agents and rounds.

\paragraph{Token Scaling.} We assess the scaling trends of token cost and accuracy under both MAD and GD through by increasing the number of rounds or agents. First, as illustrated in Figure \ref{fig_scaling_token}, with the increase in token cost, both MAD and GD exhibit an overall upward trend in accuracy. And initially the accuracy increases rapidly, but as the token cost becomes very large, the rate of accuracy growth slows down. Moreover, GD consistently outperforms MAD with scaling of tokens across all four datasets. While MAD's accuracy tends to converge as the token cost becomes exceedingly large, GD still potentially exhibits a growing trend. And we notice that GD has more sharply increasing points, which may be indicative of emergent intelligence in the token scaling of GD. It's an intriguing research point to explore scaling laws about accuracy and efficiency within multi-agent debate.

\subsection{Ablation Study}
\label{ablation-study}


\begin{figure}[t]
    \centering
    \includegraphics[width=1\columnwidth]{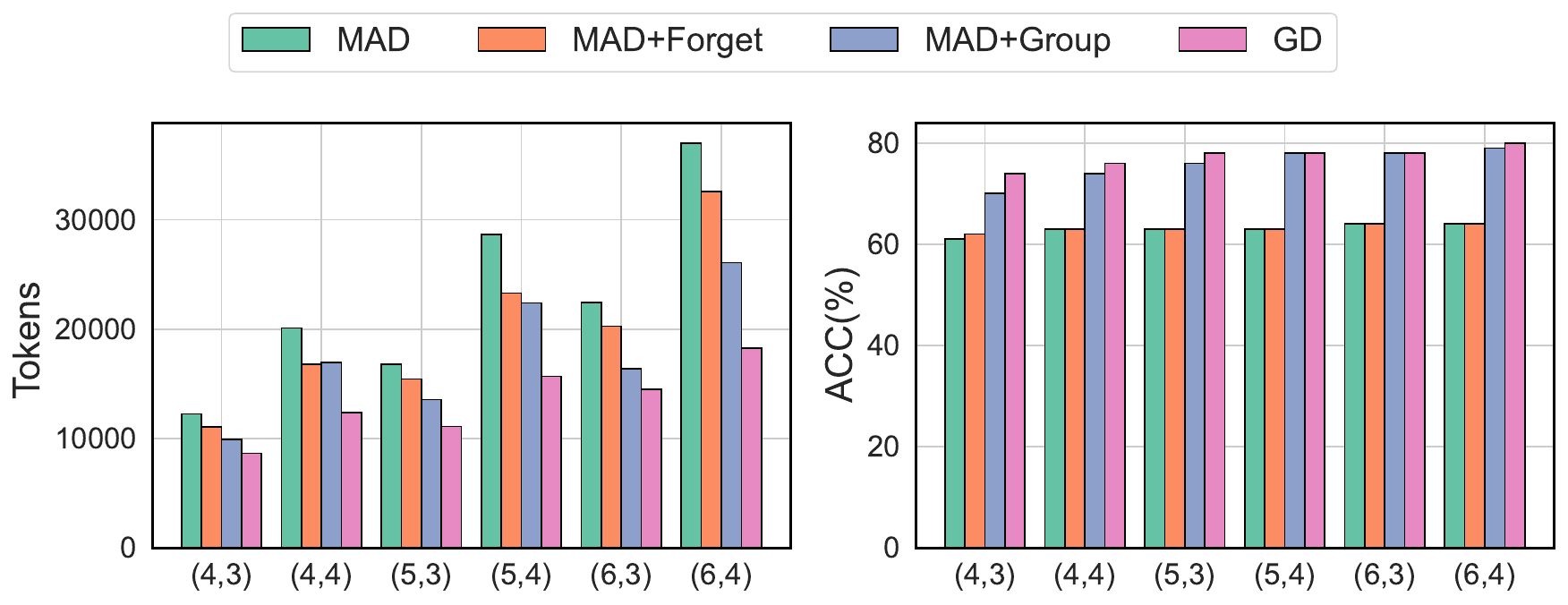}
    \caption{\textbf{Ablation Study.} }
    \label{fig9}
\end{figure}

In order to further investigate the impact of certain components in GD, we conduct a comparative analysis of MAD, MAD+Forget (MAD with only preserving summaries from the previous round), MAD+Group (MAD with group discussion) and GD. First, as illustrated in the Figure \ref{fig9}, GD outperforms all MAD and its variants in token cost and accuracy, which shows the effectiveness of involving both forget mechanism and group discussion in our method. Second, through comparing MAD+Forget with MAD and GD with MAD+Group, the forget mechanism can effectively reduce token cost while maintaining accuracy almost unchanged, which suggests that there is no need for agents to remember all summary results. Third, MAD+Group, compared to MAD+Forget, reduces a substantial number of tokens and significantly improves accuracy. This further highlights the effectiveness of our proposed group discussion method. Based on the grouping strategy analyzed previously, we hypothesize that the primary reason for the enhancement in accuracy is due to the diversity preserved among the groups.
\section{Related Work}
\label{sec:related-work}

\subsection{LLM Reasoning}

Numerous research have explored to enhance the logical reasoning capabilities of LLMs. Chain-of-Thought \cite{wei2023chainofthought} is a pioneering work that mirrors human thought processes in a step-by-step way when tackling complex problems. Self-Consistency with CoT \cite{wang2023selfconsistency} samples multiple reasoning path and selects the most consistent answer. Tree-of-Thoughts \cite{yao2024tree} allows LLMs to determine their next course of action by considering various reasoning paths and self-evaluation choices. Graph-of-Thoughts \cite{besta2024graph} further represents the nonlinear task resolution process of LLMs as an arbitrary graph and reasoning on the graph. 
Additionally, \cite{shum2023automatic} involves creating a pool of CoT candidates and selecting the optimal candidate based on certain conditions. However, these methods either use only a single agent or lack communication between agents, which makes them prone to hallucinations or self-perception errors.
Verification \cite{lightman2023lets} and feedback recording are used to enhancement reasoning capabilities. 
STaR \cite{eric2022star} generates multiple chains of thought, from which effective ones are selected.
\cite{zhou2022large} proposes a method for selecting the optimal prompt from the candidate set. 
Skeleton-of-Thought \cite{ning2023skeleton} firstly generates skeleton of answer, followed by the parallel complete of content for each point in the skeleton, thus accelerating answer generation. Table-of-Thoughts \cite{jin2023tab} enhances the accuracy of reasoning through the structured modeling of the reasoning process.

\subsection{Multi-agent Debate}

In multi-agent collaboration, the MAD approach has been demonstrated as an effective orthogonal enhancement in logical reasoning. \cite{liang2023encouraging} proposes a MAD framework that encourages divergent thinking in LLMs, where a judge manages the debate and obtain a final solution. 
\cite{du2023improving} further investigates the impact of the number of agents and rounds of debate on accuracy. \cite{xu2023toward} proposes a multi-agent collaboration strategy that simulates the academic peer review process. \cite{wang2023apollo} integrates a prior knowledge retrieval into the debate process, thereby enhancing reasoning capabilities. \cite{fu2023improving} employs autonomous enhancement of negotiation strategies using a multi-round negotiation game exploration model with two agents. \cite{chan2023chateval} presents various communication strategies and evaluates the effects of these differing approaches. Corex \cite{sun2023corex} employs collaborative methods such as debate, review, and retrieve among multiple agents. \cite{li2024improving} proposes that the utilization of a sparse communication topology in MAD to enhance performance and mitigate computational costs. Recent studies \cite{becker2025stayfocusedproblemdrift} have identified the critical issue of problem drift – a gradual deviation from the original task objective. ConfMAD \cite{lin2025enhancingmultiagentdebateperformance} integrates confidence expression throughout the debate process to improve effectiveness andperformance. 
\section{Conclusion}
\label{sec:conclusion}
In this work, we address the critical challenge of token efficiency in multi-agent debate systems, which has emerged as a key bottleneck in scaling collaborative reasoning frameworks.  We propose a novel GroupDebate method, which leverages the group discussion to mitigate this issue while fostering a diverse range of viewpoints. 
Specifically, we divide all participating agents into several debate groups, where each agent can engage in both intra-group debates and inter-group exchanges of ideas. 
Experimental results across four logical reasoning datasets demonstrate GroupDebate can significantly reduce token cost as well as enhance accuracy in multi-agent debates. 
In the future, we will further explore the theorem of how group discussion can improve accuracy and theoretically determine the optimal settings in GroupDebate. 
\section{Limitations}
\label{sec:Limitations}
While GroupDebate demonstrates significant advancements in token efficiency and reasoning accuracy, several limitations and avenues for future research remain:
The first key limitation is that we only theoretically analyze the constraints of $N$ and $S$ required to achieve optimal token cost complexity but we have not delved into the optimal settings of $N$ and $S$ and the underlying reasons why GroupDebate can potentially improve the accuracy of MAD.  However, determining the optimal values of N and S also requires considering accuracy to maximize it under the same token cost, which is very complex and necessitates more evaluations and experiments to deduce the theoretical basis for the enhancement of accuracy and optimal settings in GroupDebate.
Furthermore, although GroupDebate can significantly reduce token cost in muti-agent debates, its token cost is still higher than single-agent methods like CoT. It is necessary to explore more ways to further reduce token cost while ensuring high accuracy.


\bibliographystyle{ACM-Reference-Format} 
\bibliography{sample}

\clearpage
\appendix
\section{GroupDebate Algorithm} \label{appendix:gd-al}
\begin{breakablealgorithm}
    \caption{GroupDebate Methods}
    \label{alg:GroupDebate}
    \begin{algorithmic}[1]
        \Require Number of groups $N$, number of agents  $M$, question $Q$, total rounds $T$, intra-group debate round $R$, total stages $S$, answer extracter $VOTE$ 
        \Ensure $Answer$
     
        \State $A \gets [A_1,A_2,\ldots,A_M]$ 
        \Statex \Comment{Initialize and shuffle the agents randomly}

        \State $G \gets [G_1,G_2,\ldots,G_N]$  \Statex \Comment{Initialize each group} 

        \State $H \gets [H_1,H_2,\ldots,H_M]$  \Statex \Comment{Initialize each agent with empty memory}
        
        \State $Summary \gets [Summary_1,$
        \Statex $\qquad \qquad Summary_2,\ldots,Summary_N]$  
        \Statex \Comment{Initialize summary pool with empty list}

        \For{$i=1$ to $M$}
            \State $H_i \gets [Q]$ \Comment{Initialize memory of each agent}
        \EndFor
        
        \For{$s=1$ to $S$}
            \For{$j=1$ to $N$}
                \For{$t = (s-1)R+1$ to $\min(sR,T)$}
                    \For{$A_i \in G_j$}
                        \If{$s=1$ \textbf{and} $t=1$}
                            \State $h_i \gets A_i(H_i)$ 
                            \State $H_i \gets H_i + h_i$ 
                            \State $H_i \gets H_i + BUF$ 
                        \ElsIf{$s \neq 1$ \textbf{and} $t=(s-1)R+1$} 
                            \State $h_i \gets A_i(H_i)$ 
                            \State $H_i[-2] \gets h_i$    
                        \Else
                        \For{$A_{i'} \in G_j$ \textbf{and} $A_{i'} \neq A_i$}
                            \State $buf \gets [\quad]$ 
                            \State $buf \gets buf + Replay_{i'}$  
                        \EndFor
                        \State $H_i[-1] \gets buf$ 
                    
                        \State $h_i \gets A_i(H_i)$  
                        \State $H_i[-2] \gets h_i$   
                        \EndIf
                    \EndFor
                \EndFor
                \If{$s \neq S$}
                    \State $summary \gets [\;]$
                    \For{$A_i \in G_j$}                       
                        \State $summary \gets summary + H_i[-2]$             
                    \EndFor
                    \State $Summary_j \gets$
                    \Statex $\qquad \qquad \qquad \qquad \qquad LLM(summary)$ 
                    
                \EndIf 
            \EndFor
            \For{$i=1$ to $M$}
                \State $H_i[-1] \gets Summary $
            \EndFor

        \EndFor
        \State $Answer \gets VOTE(H)$
        
    \State \Return $Answer$
    \end{algorithmic}
\end{breakablealgorithm}

\section{Token Cost Analysis} \label{appendix:token-cost-analysis}
In this appendix section, we aim to provide a theoretical analysis of the token cost For both MAD and GD. As LLMs' outputs typically are not too long and we can actually control the token length of LLMs' outputs in prompts to some extent, we assume that the upper bound on the number of tokens output by each agent participating in debate is $Output_{max}$ and the upper bound on the number of tokens in the generated summary is $Summary_{max}$. We define $C$ as the maximum of $Output_{max}$ and $Summary_{max}$.
\subsection{Token Cost in MAD} \label{appendix:token-cost-mad}
Here, we implement the MAD method, which summarizes the responses from other agents and inputs all previous summaries For each agent in each round. The token cost includes both input and output cost, and in each round $t$, it can be divided into two parts: summary generation $Token^{summary}_{t}$ and agents' responses $Token^t$. Thus, the total token cost $Token^{MAD}$ can be represented as:
\begin{equation}
\begin{aligned}
Token^{MAD} = Token^{1} + \sum_{t=2}^{T} (Token^{summary}_{t-1}\\
+Token^{t})
\end{aligned}
\end{equation}
SpecIfically, we provide a detailed description of the token cost For each part. (1) \textbf{summary generation}: The token cost For each agent includes the output from other agents and output summary. (2) \textbf{agents' responses}: If $t=1$, the token cost For each agent includes the initial question prompt and its own output. If $t>1$, the token cost For each agent includes the current summary, its own output, and the total token cost of all its previous inputs and outputs. The detailed computation process of the token cost in MAD can be found in Algorithm \ref{alg:token_cost_mad}.
\begin{algorithm}[t]
    \caption{Token Cost in MAD Methods}
    \label{alg:token_cost_mad}
    \begin{algorithmic}[1]
        \Require Number of groups $N$, number of agents  $M$, question length $Q$ , total rounds $T$,  output length of each agent $A_i(i=1,2, \ldots,M)$ in each round $t(t=1,2,\ldots,T)$ $Output_i^t$, the summary of the output without $A_i$ in each round $t(t=1,2,\ldots,T-1)$ $Summary_i^t$
        \Ensure Total token cost $Token^{MAD}$
        
        \State $Token^{1} \gets M \times Q + \sum_{i=1}^{M} Output_i^1$ 
        \Statex \Comment{First round}
        
        \For{$t = 2$ to $T$}
            \State $Token^{summary}_{t-1} \gets \sum_{i=1}^{M}(\sum_{i' \neq i}$ 
            \Statex $\qquad \qquad Output_{i'}^{t-1} +  Summary_{i}^{t-1})$ 
            \Statex \Comment{Summary stage}
            
            \State $Token^{t} \gets Token^{t-1}+\sum_{i=1}^{M}($
            \Statex $\qquad \qquad Summary_{i}^{t-1} + Output_{i}^{t})$ 
            
            \State $Token^{t} \gets \sum_{i=1}^{M}(\sum_{t'=1}^{t-1} (Output_i^{t'}$
            \Statex $\qquad\qquad  +Summary_i^{t'})+ Q + Output_{i}^{t})$ 
            \Statex \Comment{Subsequent rounds}
        \EndFor
        \State $Token^{MAD} \gets Token^{1} + \sum_{t=2}^{T} ($
        \Statex $ \qquad Token^{summary}_{t-1}+Token^{t})$
        \Statex \Comment{Total token cost in debate}
        
        \State \Return $Token^{MAD}$
    \end{algorithmic}
\end{algorithm}

Following the line 7 in Algorithm \ref{alg:token_cost_mad}, with $Output_i^{t} \leq Output_{max}$ and $Summary_i^{t} \leq Summary_{max}$ For every $t$ and $i$, we can infer the following:
\begin{equation}
\begin{aligned} 
    &Token^{MAD} = \\
    &\quad MTQ + \sum_{t=1}^{T}\sum_{i=1}^{M}Output_i^t + \sum_{i=1}^{M}\sum_{t=2}^{T}(\sum_{i' \neq i} \\
    & \quad Output_{i'}^{t-1} +Summary_i^{t-1})\\ 
    & \quad + \sum_{i=1}^{M}\sum_{t=2}^{T}\sum_{t'=1}^{t-1}(Output_i^{t'}+Summary_i^{t'}) \\ 
    & < MTQ + (M^2T+\frac{1}{2}MT^2)\times Output_{max} \\
    & \quad + (\frac{1}{2}MT^2+\frac{1}{2}MT)\times Summary_{max}
\end{aligned}
\end{equation}
ThereFore, we can finally obtain $Token^{MAD} = \mathcal{O}\left(MTQ+(M^2T+MT^2)C\right)$.

\subsection{Token Cost in GroupDebate} \label{appendix:token-cost-gd}
As mentioned in Section \ref{multi-agent-group-debate}, our GroupDebate includes three types of processes and thus the total token cost $Token^{GD}$ can be further dividied into:
\begin{equation}
\begin{aligned} 
    &Token^{GD} = \underbrace{Token_1^1}_\text{initial thinking} + \\ 
    &\underbrace{\sum_{s=2}^{S}(Token^{summary}_{s-1}+Token_s^{(s-1)R+1})}_\text{inter-group debate} \\
    &+ \underbrace{\sum_{s=1}^{S}\sum_{t=(s-1)R+2}^{min(sR,T)}Token_s^t}_\text{intra-group debate}
\end{aligned}
\end{equation}
\label{eq:gd_divide}
SpecIfically, For initial thinking, the token cost of each agent includes the initial question prompt and its own output. For intra-group debate, the token cost of each agent includes all responses from other agents within the same group in the previous round and its output. For inter-group debate, the token cost of each agent includes the summary generation cost, which comprises the responses from other groups and the output summary, as well as its own output. The detailed computation process of the token cost in GroupDebate can be found in Algorithm \ref{alg:token_cost_gd}.

\begin{algorithm}[ht]
    \caption{Tokens Cost in GroupDebate Methods}
    \label{alg:token_cost_gd}
    \begin{algorithmic}[1]
        \Require Number of groups $N$, number of agents  $M$, question length $Q$, total rounds $T$, group debate round $R$, total stages $S$, summary of each group at the end of each stage $Summary = \{Summary_j^s|j=1,2, \ldots,N,s=1,2,\ldots,S\}$ , output length of each agent $A_i(i=1,2, \ldots,M)$ in each round $t(t=1,2,\ldots,T)$ $Output_i^t$,  each group agents set $G = \{G_j|j=1,2, \ldots,N\}$
        \Ensure Total token cost $Token^{GD}$
        
        \State $Token_1^{1} \gets M \times Q + \sum_{i=1}^{M} Output_i^1$ 
        \Statex \Comment{First round} 
        
        \For{$t = 2$ to $R$}
            \State $Token_1^t \gets \sum_{j=1}^{N} \sum_{i \in G_j} (Q  + Output_i^{t}+$
            \Statex $\qquad \qquad \sum_{i' \in G_j} Output_{i'}^{t-1})$ 
            \Statex \Comment{Subsequent rounds of the first stage}
        \EndFor
        
        \For{$s=2$ to $S$}
            \State $Token_{s-1}^{summary} \gets \sum_{j=1}^{N} (\sum_{i \in G_j}$
            \Statex $\qquad \qquad Output_i^{(s-1)R} + Summary_j^{s-1})$ 
            \Statex \Comment{Summary at the end of stage $s-1$}
        
            \State $Token_s^{(s-1)R+1} \gets \sum_{i=1}^{M} (Q+$
            \Statex $\qquad \quad Output_i^{(s-1)R}+ \sum_{j=1}^{N} Summary_j^{s-1}$
            \Statex $\qquad \quad + Output_i^{(s-1)R+1})$ 
            \Statex \Comment{First round of the stage $s$}
        
            \For{$t = (s-1)R+2$ to $\min(sR,T)$}
                \State $Token_s^t \gets \sum_{j=1}^{N} \sum_{i \in G_j} (Q  + $
                \Statex $\qquad \qquad Output_i^{t} + \sum_{i' \in G_j} Output_{i'}^{t-1})$ 
                \Statex \Comment{Subsequent rounds of the stage $s$}
            \EndFor
        \EndFor
        
        \State $Token^{GD} \gets  \sum_{t=1}^{R}Token_1^t+$
        \Statex $\qquad \qquad \sum_{s=2}^S(Token_{s-1}^{summary}+$
        \Statex $\qquad \qquad \sum_{t=(s-1)R+1}^{\min(sR,T)}Token_s^t)$ 
        \Statex \Comment{Total token cost in debate}
        
        \State \Return $Token^{GD}$
    \end{algorithmic}
\end{algorithm}

Following Appendix~\ref{appendix:token-cost-mad} and Eq.~\ref{eq:gd_divide}, we have: 
\begin{equation}
\begin{aligned}
    &Token^{GD}= \\
    &\quad MQ + \sum_{i=1}^M Output_i^1 + \sum_{s=2}^S [\sum_{j=1}^N(\sum_{i\in G_j} \\ 
    & \quad Output_i^{(s-1)R}+Summary_j^{s-1}) + \\
    & \quad \sum_{i=1}^M(Q+Output_i^{(s-1)R}+\sum_{j=1}^N Summary_j^{s-1} +\\
    & \quad Output_i^{(s-1)R+1})] + \sum_{s=1}^S \sum_{t=(s-1)R+2}^{min(sR,T)}\sum_{j=1}^N\sum_{i\in G_j}\\
    &\quad (Q+Output_i^t+\sum_{i' \in G_j}Output_{i'}^{t-1}) \\
    & \leq MTQ + [3MS-2M+(T-S)(K+1)M]\\
    & \quad \times Output_{max} + (S-1)(M+1)N\\
    & \quad \times Summary_{max} \\
    & \leq MTQ+\frac{2M^2T}{N}\times Output_{max} + \\
    & \qquad 2MSN\times Summary_{max} \\
    & = \mathcal{O}\left(MTQ+(\frac{M^2T}{N}+MSN)C\right)
\end{aligned}
\end{equation}

When we set $N \rightarrow \mathcal{O}\left(\sqrt{\frac{MT}{S}}\right)$, we can theoretically obtain $Token^{GD} \rightarrow \mathcal{O}\left(MTQ+\sqrt{M^3TS}C\right)$. Furthermore, If we consider setting $S$ to a very small positive integer, then $Token^{GD}$ can approach $\mathcal{O}\left(MTQ+\sqrt{M^3T}C\right)$. This complexity is signIficantly lower than that of MAD.

\section{Prompts} \label{appendix:prompts}
In this section, we present some examples of prompts. Table \ref{tab:input-prompts} displays the input prompts used in our GroupDebate across dIfferent datasets, which encompass five dIfferent types. Table \ref{tab:Requirements} outlines the prompts regarding output Format Requirements in our GroupDebate.

\begin{table*}[t]
\setlength{\tabcolsep}{5.5pt}
\centering
\resizebox{\textwidth}{!}{
\begin{tabular}{ccl}
     {\bf Type} & Task & {\bf Prompt} \\
      \midrule
     \multirow{4}{*}{System}  & \multirow{4}{*}{All} & \emph{Welcome to the debate! You are a seasoned debater with expertise in succinctly and persuasively expressing your viewpoints. }\\
     & & \emph{You will be assigned to debate groups, where you will engage in discussions with fellow participants. The outcomes of }\\
     & & \emph{each group's deliberations will be shared among all members. It is crucial For you to leverage this inFormation effectively }\\
     & & \emph{in order to critically analyze the question at hand and ultimately arrive at the correct answer. Best of luck!} \\
     \midrule
      \multirow{4}{*}{Starting}  & \multirow{1}{*}{Arithmetic} & \emph{What is the result of \{\}+\{\}*\{\}+\{\}-\{\}*\{\}? <Output Format>.}   \\
      \cmidrule{2-3}
      & \multirow{1}{*}{GSM8K} & \emph{Can you solve the following math problem? <Problem> Explain your reasoning. <Output Format>.}\\
      \cmidrule{2-3}
     & \multirow{1}{*}{MMLU} & \emph{Can you answer the following question as accurately as possible? {}: A) {}, B) {}, C) {}, D) {} Explain your answer, <Output Format>.} \\
     \cmidrule{2-3}
     & \multirow{1}{*}{MATH} & \emph{Can you solve the following math problem? <Problem> Explain your reasoning as concise as possible.<Output Format>. } \\
     \midrule
      \multirow{2}{*}{Intra-group Debate}  & \multirow{2}{*}{All} & \emph{These are the recent opinions from other agents: <other agent responses> Using the opinions} \\
      & & \emph{carefully as additional advice, can you provide an updated answer?}\\
      && \emph{Examine your solution and that other agents step by step. <Output Format>.} \\
     \midrule
     \multirow{3}{*}{Summary}  & \multirow{3}{*}{All} & \emph{These are the recent/updated opinions from all agents: <all agent responses>}\\
     & & \emph{Summarize these opinions carefully and completly in no more than 80 words. }\\
     & & \emph{Aggregate and put your final answers in parentheses at the end of your response. }\\
     \midrule
     \multirow{3}{*}{Inter-group Debate}  & \multirow{3}{*}{All} & \emph{These are the recent opinions from all groups: Your group response: < group summary>, Other group responses: } \\
      & & \emph{<other group summary>. Using the reasoning from all groups as additional advice, can you give an updated answer?} \\
      & & \emph{Examine your solution and that all groups step by step. <Output Format>.} \\
    \bottomrule
\end{tabular}

}
\caption{\textbf{Prompts in Each Stage.} List of prompts used in each task. }
\label{tab:input-prompts}
\end{table*}

\begin{table*}[t]
\centering
\resizebox{\textwidth}{!}{
\begin{tabular}{@{\extracolsep{2pt}}lc@{}}
\multicolumn{1}{c}{Dataset} & \multicolumn{1}{c}{Output Format Requirements} \\
\midrule
\multirow{1}{*}{Arithmetic} & \emph{Make sure to State your answer at the end of the response.}\\
\midrule
\multirow{2}{*}{GSM8K} & \emph{Your final answer should be a single numerical number, in the Form \textbackslash boxed\{\{answer\}\},}\\
& \emph{at the end of your response.}\\
\midrule
\multirow{1}{*}{MMLU} & \emph{Put your final choice in parentheses at the end of your response.}\\
\midrule
\multirow{1}{*}{MATH} & \emph{Put your final answer in the Form \textbackslash boxed\{\{answer\}\}, at the end of your response.}\\
\bottomrule
\end{tabular}
}
\caption{\textbf{Output Format Requirements in Each Dataset.}}
\label{tab:Requirements}
\end{table*}

\newpage


\end{document}